# Automated Estimation of Construction Equipment Emission using Inertial Sensors and Machine Learning Models

**Farid Shahnavaz[a] and Reza Akhavian[b*]**

[a] Graduate Student, Department of Civil, Construction, and Environmental Engineering, San Diego State University, San Diego, CA 92182, United States of America

[b] Assistant Professor, Department of Civil, Construction, and Environmental Engineering, San Diego State University, San Diego, CA 92182, United States of America, e-mail: rakhavian@sdsu.edu

[*] Corresponding Author

**Abstract**

The construction industry is one of the main producers of greenhouse gasses (GHG). Quantifying the amount of air pollutants including GHG emissions during a construction project has become an additional project objective to traditional metrics such as time, cost, and safety in many parts of the world. A major contributor to air pollution during construction is the use of heavy equipment and thus their efficient operation and management can substantially reduce the harm to the environment. Although the on-road vehicle emission prediction is a widely researched topic, construction equipment emission measurement and reduction have received very little attention. This paper describes the development and deployment of a novel framework that uses machine learning (ML) methods to predict the level of emissions from heavy construction equipment monitored via an Internet of Things (IoT) system comprised of accelerometer and gyroscope sensors. The developed framework was validated using an excavator performing real-world construction work. A portable emission measurement system (PEMS) was employed along with the inertial sensors to record data including the amount of $CO$, $NO_X$, $CO_2$, $SO_2$, and $CH_4$ pollutions emitted by the equipment. Different ML algorithms were developed and compared to identify the best model to predict emission levels from inertial sensors data. The results showed that Random Forest with the coefficient of determination ($R^2$) of 0.94, 0.91 and 0.94 for $CO$, $NO_X$, $CO_2$, respectively was the best algorithm among different models evaluated in this study.

## 1. Introduction

Global warming is one of the most serious challenges facing humankind today. A rise in the concentration of greenhouse gases (GHGs) in the atmosphere, such as $CO_2$, $CH_4$, $NO_2$, and water vapor, is responsible for the increase in average earth surface temperature and $CO_2$ plays a key role in contributing to this temperature growth [1, 2]. Since 1976, human activities have been the main cause of global warming, which in turn has many negative effects on climate change, agriculture, and human health [3-6]. The construction industry produces 23% of the total $CO_2$ emissions released by human activities [7]. Construction materials production, transportation of building modules, and the large amount of construction equipment fuel consumption are all responsible for GHG emissions [8-10] . Construction equipment typically consumes a large amount of energy and contributes significantly to GHG emissions and global warming. Yan et al. showed that 12-17% of the total GHG emissions in building construction are from the transportation of building materials and construction equipment energy consumption [11]. In 2005, there were more than two million pieces of construction and mining heavy equipment in the United States consuming more than 6 billion gallons of diesel fuel annually [12], a figure that most certainly has not decreased ever since. Thus, efficient management and use of construction equipment have a great effect on global warming and earth temperature. Heavy equipment often uses powerful engines and produces different types of gasses even during idling. Accurate field measurement of these emissions is the first step in understanding and management of energy consumption. Considering the significant uncertainties that exist in inventories of emissions from construction equipment, accurate methods are required to estimate the exact amount of emissions [13]. Currently, such accurate measurements are manual, time-consuming, and labor-intensive [14].

This paper describes the development of a novel framework that uses machine learning (ML) methods to predict the emissions of heavy construction equipment using data collected by the accelerometer and gyroscope sensors and the validation of this framework using an excavator performing real-world construction work. Emission was measured using a portable emission measurement system (PEMS) with a probe inserted into the exhaust pipe of the equipment and recorded the number of $CO$, $NO_X$, $CO_2$, $SO_2$ and $CH_4$ in ppm. The PEMS device allows measuring the emissions of combustion engine vehicles and equipment while they are being operated instead of those used only on stationary rollers on a dynamometer that simulates real-world driving [15]. Previous studies have largely focused on predicting emissions from on-road vehicles, turbines, and diesel engines leveraging engine features and direct measurements [16-20]. However, to the best of the authors' knowledge, this is the first research study that enables ML-based emission estimation of construction equipment using non-intrusive sensing of the equipment movement with no regard to the machine engine, fuel consumption, and or speed. In this research, the movements and the vibration of the equipment articulated parts are used as the inputs of the ML model and the outputs are the amounts of emissions produced by the equipment. The predictive performance of four different supervised learning algorithms are evaluated and compared and the best algorithm that can predict the output in each case is introduced.

The contribution of this study to the body of knowledge and practice is the demonstration of the fact that activity-based emission estimation can be made practical and easy and be performed while the equipment is engaged in actual work without the need to sample from the equipment exhaust pipe. This investigation has been inspired by previous studies of the second author where activities of construction equipment were successfully detected with a promising accuracy using similar methods. Therefore, the objective of this study was to determine whether an inertial sensor-based ML model can be adopted to quantify the amount of emission since different equipment activities require different levels of engine engagement and thus fuel consumption. The rest of this paper is organized as follows. First, a comprehensive review of the literature published on the topics of off- and on-road equipment emission estimation is presented. Next, the proposed methodology including a description of the real-world data collection session is provided. After that, the results of implementing the proposed methodology are described and a discussion of the results as well as a summary and conclusion of the research is presented.

## 2. Literature Review

The literature on transportation emissions generated by combustion engines can be divided into studies that focused on on-road and those that studied off-road vehicles. Each category of research is characterized by unique features, processes, standards, and limitations that deserve a full review to establish the state of the research. In addition, the use of IoT methods with accelerometer and gyroscope sensors for indirect measurement of different phenomena or physical properties have recently gained significant research attention. Therefore, this section provides a comprehensive literature review in each of these three domains: (a) prediction of the emissions from on-road vehicles; (b) prediction of the emissions from construction (i.e., off-road) equipment; and (c) the use of accelerometer and gyroscope data to train ML models for estimation or prediction purposes.

### 2.1. On-road Vehicles

The emission performance of on-road combustion engine vehicles is influenced by the size and type of the engine, fuel type, and exhaust after-treatment system used [21]. Most of the studies in this area concentrated on particulate emissions [22] including $NO_X$ emissions [23-28]. Some studies used ML methods to predict the emissions, using dynamometer tests and the parameters to develop these models. Si et al. developed a model to predict $NO_X$ emissions using a neural network (NN) where the actual values were measured by a Continuous Emissions Monitoring System (CEMS) [29] (PEMS, which is the device used in the study presented in this paper, was designed to provide an alternative to the drawbacks of the CEMS [30].) Built upon the findings of their first study, Si et al. incorporated gradient boosting techniques in subsequent research to improve their results [31]. Other researchers have shown that the combination of neural networks and heuristic algorithms would further enhance the results further [15]. In another study, multiple factors such as road environment, atmospheric, and after-treatment performance were considered when analyzing $NO_X$ concentrations from Euro 6 diesel engines during real-world driving experiments [32]. Wen et al. considered data such as vehicle speed, vehicle acceleration, and exhaust gas recirculation (EGR) as input to train a NN nonlinear autoregressive exogenous model (NARX) [33]. Khurana et al. reviewed different supervised learning algorithms to predict emissions from automobiles and concluded that NN had accurate answers in addressing this kind of problem [34]. Fei et al. used a classification model based on the CatBoost algorithm to categorize emission levels [35]. Le-Cornec et al. clustered vehicles with similar emissions performance then modeled instantaneous emissions [36]. In a more recent study, Yu et al.

developed a deep learning algorithm to predict the instantaneous $NOx$ emissions from diesel engines [37].

Previous studies in this area have focused on on-road vehicles and have demonstrated promising results. Nevertheless, measuring and predicting on-road vehicles emission pose fewer challenges compared to off-road equipment. The diverse set of activities and operations of off-road vehicles require a variety of engine modes and thus fuel consumption and emission levels. In addition, the uncertainty of such activities adds to the level of complexity and challenges involved in developing predictive models of any sort.

*2.2. Off-road Vehicles*

Compared to emission estimation research for on-road vehicles, off-road vehicles have received much less attention. The dearth of research is even more severe concerning using advanced technologies such as ML methods for emission prediction. A number of studies conducted by federal organizations such as the Environmental Protection Agency (EPA), academic institutions, and private organizations (e.g., Clean Air Technologies International, Inc. (CATI)), used onboard instruments to measure construction equipment emission [38-41]. While the insights provided by these studies inspire new research, it has been reported that in none of these cases, collected data are verified for quality or made available for public use [42]. Heidari and Marr compared real-time emissions from construction equipment with model predictions proposed by EPA [14]. Their findings indicate that although model predictions agreed with actual emissions in some cases, in others they were up to 100 times higher. In addition, they obtained very different emission rates during various operating conditions. Abolhasani et al. assessed the fuel use and emissions of excavators during field duty cycles and concluded that in non-idle modes, mass per time emission rates were 7 times higher than in idle modes.[43]. Data collected on the emissions of backhoes, motor graders, and wheel loaders using B20 biodiesel and petroleum diesel was compared by Frey et al. where a methodology for designing the study, collecting field data, screening and ensuring the quality of the data, and analyzing the data was developed. They showed that using B20 instead of petroleum diesel would lead to a 1.8% insignificant decline in nitric oxide (NO) emissions, as well as significant decreases in opacity, hydrocarbons (HC), and carbon monoxide (CO) emissions, respectively [44]. A study of selected motor graders fueled by petroleum diesel and B20 biodiesel characterized their field activity, fuel use, and emissions and concluded that using B20 instead of petroleum diesel results in a negligible decrease in emissions. [45]. Lewis et al. developed recommendations for reducing the emissions from construction equipment and recommended using field emissions data instead of engine dynamometer data to reduce emissions [46]. In another study, the development and use of an emissions inventory system for a fleet of backhoes, front-end loaders, and motor graders were discussed to support the decision-making process regarding the replacement of older equipment with more efficient ones [47]. Using engine dynamometer data from nonroad mobile sources such as construction, farming, and industrial engines, Frey and Bammi developed probabilistic emission factors for $NO_X$ and $HC$ [48]. Frey et al. did a sensitivity analysis to predict fuel consumption and emissions for construction equipment through engine attribute data including horsepower, displacement, model year, engine tier, and engine load and showed that in petroleum diesel engines, fuel use and pollutant emission rates increase with gear ratio, horsepower, and torque and decrease with model year and engine tier [49]. By analyzing field data, a quantitative model was developed by Barati and Shen

(2016) to more accurately estimate the various emission rates of construction equipment [50]. An operational level emission model has been developed based on ordinary least square (OLS) and multivariate linear regression (MLR) analyses of field data. The results of that study verified the high correlation between emission rates and operational parameters and engine data.

As stated earlier, there is a general scarcity of research in the area of off-road vehicle emission estimation. Of the limited studies in this area, a small subset focused on construction equipment. This is while the understanding of construction heavy machinery emission is critical to prevent GHG emission as well as to support the decision-making process in bidding and other project phases [51]. Standards such as those suggested by the EPA and other academic research do not represent an adequate level of prediction accuracy and they do not consider different operating states. In addition, to the best of the authors' knowledge, none of the previous studies in this area leveraged the potential of movement-based methods such as using accelerometer and gyroscope sensors in conjunction with advanced ML methods.

*2.3. Training ML Models with Inertial Data*

There are generally two types of inertial measurement systems that are pervasively used and commonly found in daily lives (such as in smartphones): a sensor that measures acceleration called an accelerometer, and a sensor that measures the velocity of rotation about a circular axis (i.e., angular velocity) called gyroscope [52]. A magnetometer could be added to the two sensors in which case they are collectively called an inertial measurement unit (IMU). Researchers have used wearable devices equipped with one or two of these sensors or a complete IMU to report instantaneous and sudden vibrations of the human body since the 1990s [53-56]. Micro-Electro-Mechanical Systems (MEMS) inertial sensors have proved extremely useful, accurate, and computationally efficient for activity recognition with applications in health care, sports, and engineering [57-59]. For instance, researchers analyzed soccer players' movement patterns using wireless accelerometers to gain insight into pattern recognition [60]. With the advancement of smartphones, mobile accelerometer sensors were later used to identify human activity [61]. Motoi et al. proposed taking measurements of the speed at which the subjects walk and monitoring their posture and movement [62]. Combined with a body-worn microphone, accelerometers were used in a wood workshop to segment and recognize typical user gestures [63]. Similar research has recently gained significant traction in construction engineering and management applications for workers and equipment activity recognition [64, 65]. For example, Johnson and Trivedi used accelerometers to detect, recognize, and record driving styles for a driving safety application [66]. Using different types of sensors, researchers have also explored simulation-based heavy equipment emission estimation [67]. The use of ML models trained by the accelerometer and gyroscope data has been explored as well. Akhavian and Behzadan (2015) developed a ML-based methodology to use smartphone sensors as ubiquitous, multi-modal data collection and transmission nodes to detect detailed construction workers [68] and equipment [69] activities. More recently and with the advancement of deep learning, Salton et al. (2021) developed a framework for recognizing heavy construction equipment activity via accelerometers based on convolutional and recurrent neural network architectures [70].

As shown in this part, the concentration of previous studies using inertial sensors was focused on activity recognition in humans and equipment. Previous studies have often developed a classification model to differentiate between several defined activities. Frameworks that use ML models trained with inertial sensors have never been adopted before to seamlessly predict emissions from heavy equipment.

## 3. Methodology

In this section, the data collection method is outlined first and then a description of the feature extraction process is provided. Next, the learning algorithms developed as well as metrics to evaluate their performance are discussed.

*3.1. Data Collection*

To generate high-fidelity outputs, the data collection process was performed in an uncontrolled environment where the construction equipment was involved in routine daily activities without any interruptions from the researchers. Data was collected from a *Caterpillar 305D CR* excavator which was working to expand a trench around an underground pipeline as shown in Figure 1. Two Noraxon MyoMotion accelerometer and gyroscope integrated sensors [71] were used to collect the equipment body acceleration (i.e., vibration) data. One sensor (Sensor 1) was attached inside the excavator cabin and another one was affixed to its stick (Sensor 2), near the bucket. In addition, the data regarding engine emissions generated by the equipment was logged by an *E-Instruments E9000 Plus Gas Analyzer* PEMS [72]. Table 1 shows the sensors' specifications [71].

*Table 1. Specifications of the accelerometer and gyroscope sensors used in the experiment.*

| | |
|---|---|
| **Channel** | Up to 16 sensors |
| **Static accuracy** | ±0.4° |
| **Dynamic accuracy** | ±1.2° |
| **Sampling frequency** | 100 Hz |
| **Data output** | Joint angles, acceleration, rotation quaternions |
| **Maximum output rate** | 400 Hz |
| **Orientation angel frequency** | 0.25 degree (pitch/roll); 1.25 degrees (heading) |
| **Anatomical angel frequency** | +/- 1.0 degree (static); +/- 2.0 degrees (dynamic) |
| **Angular velocity (Gyroscope)** | +/- 7000 degrees/sec; Internal Sampling Rate 1600Hz |
| **Acceleration (Accelerometer)** | +/- 200 g; Internal Sampling Rate 1600 Hz |
| **Motion sensor dimensions** | 1.75" L × 1.3" W × 0.48" H (4.45 cm L × 3.3 cm W × 1.22 cm H) |
| **Weight** | Less than 0.67oz (19g) |

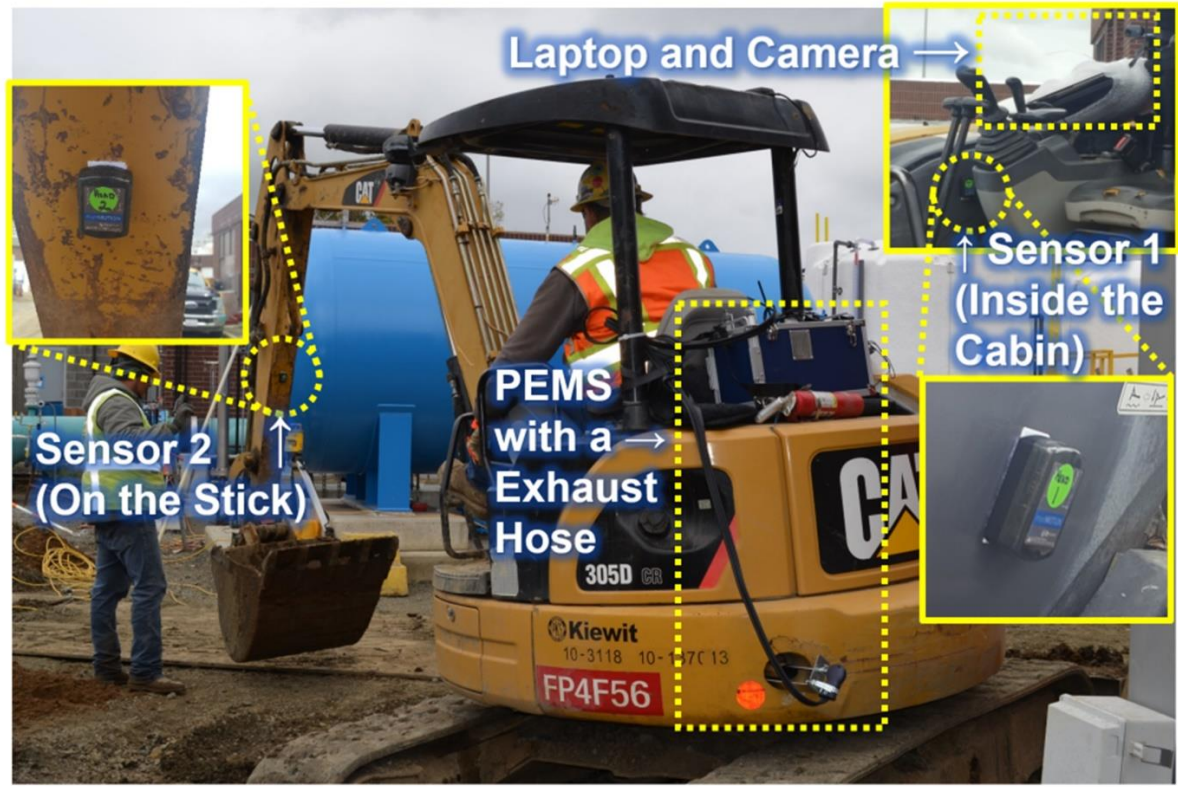

*Figure 1. Data collection setup with IoT sensors*

The accelerometer data was logged using the sensors' *Noraxon myoRESEARCH Software* and stored in a laptop secured inside the equipment cabin during data collection to maintain wireless connectivity. A camera was also attached inside the cabin to record the entire data collection session. The laptop was plugged into a portable 200 Watt *Caterpillar Power Station* to ensure it is powered on throughout data collection. The sampling frequency was set at 100 Hz. This frequency ensured that no noteworthy movement was neglected and at the same time, the volume of recorded data was not restrictively expansive. Data was stored in comma-separated value (CSV) format for further pre-processing. As shown in Table 2, the PEMS recorded *NO, $NO_2$, $CO_2$, CO, $O_2$, $SO_2$, $CH_4$*, and *$H_2S$ as* well as *Pressure* and *Temperature* [72]. PEMS uses electrochemical sensors for most gasses. Electrochemical sensors detect interactions between the sensing surface and the analytes and convert them into quantitative and qualitative evidence by using electrodes [73]. The other sensor is nondispersive infrared (NDIR) and carbon dioxide is typically measured with this type of sensor. In these sensors, light waves are emitted from an infrared (IR) lamp through a tube filled with air toward an optical filter in front of an IR light detector. Then, infrared detectors measure the light that isn't absorbed by $CO_2$ molecules or the optical filter [74]. In Table 2, $T_{air}$ and $T_{gas}$ are the temperatures of the air and gas respectively. This study focuses on *NO, $NO_2$, $CO_2$,* and *CO* emissions as major diesel engine pollutants [14]. Furthermore, the sum of *NO* and *$NO_2$* measured emission will be used as *$NO_x$* to enable comparison between the results of this study and previous work that measured this pollutant.

*Table 2. PEMS specifications.*

| Parameter | Sensor | Range | Resolution | Accuracy |
|---|---|---|---|---|
| $NO$ | Electrochemical | 0 - 5000 ppm | 1 ppm | ±5 ppm <100 ppm<br>±5 % rdg for >100 ppm |
| $NO_2$ | Electrochemical | 0 - 1000 ppm | 1 ppm | ±5 ppm <100 ppm<br>±5 % rdg for >100 ppm |
| $CO_2$ | NDIR | 0 - 50.0 % | 0.1 % | ±3 % rdg <8 %<br>±5 % rdg <50 % |
| $O_2$ | Electrochemical | 0 - 25 % | 0.1 % | ±0.2 % vol |
| $SO_2$ | Electrochemical | 0 - 5000 ppm | 1 ppm | ±5 ppm <100 ppm<br>±5 % rdg for >100 ppm |
| $CH_4$ | NDIR | 0 - 50,000 ppm | 1 ppm | ±50 ppm <2,500 ppm<br>±2 % >2,500 ppm |
| $H_2S$ | Electrochemical | 0 - 500.0 ppm | 0.1 ppm | ±5 ppm <125 ppm<br>±4 % rdg for <500ppm |
| $T_{air}$ | Pt100 | -4 to 248 °F<br>-20 to 120 °C | 0.1 °F<br>0.1 °C | ±1 °F<br>±1 °C |
| $T_{gas}$ | Tc K | -4 to 2280 °F<br>-20 to 1250 °C | 0.1 °F<br>0.1 °C | ±2 °F<br>±2 °C |

*3.2. Feature Extraction*

In supervised ML algorithms, training the models with raw data not only increases computational cost (particularly in the case of high dimensional, high volume data) but also may lead to an accuracy drop due to overfitting [75]. Therefore, extracting certain features serves as an important pre-processing step before model training. These features often include statistically derived values such as mean, variance, peak, interquartile range (IQR), correlation, and root mean square (RMS). Training ML models with accelerometer data has been subject to a great deal of previous research as stated in the Literature Review section. One of such studies conducted by the last author for construction equipment activity recognition leveraged ReliefF and Correlation-based Feature Selection (CFS) to select distinguishing features [69]. Inspired by the findings of that study, the same subset of features including the average of the rotation sensor in z and x-direction, the average of the accelerometer sensor in x, y, and z-direction, the interquartile range of the accelerometer sensor in the x-direction and the peak of the accelerometer sensor in x-direction was deployed in this research. Features are extracted from each data segment created by breaking down the dataset into windows of equal size. Typically, a 50% overlap between the segments is considered to ensure continuity in capturing all patterns in the data. A window size of 0.25s (25 data points per segment) was used based on the research conducted by Akhavian and

Behzadan (2015) where the researchers concluded that this is the optimal size for the segments in the accelerometer and the gyroscope data for construction equipment activity recognition [69].

### 3.3. Learning Algorithms and Performance Metrics

This study aims to develop different types of ML models and evaluate their performance in predicting the level of emissions of certain pollutants produced by construction equipment. Data extracted from accelerometers and gyroscopes are used as input to the model and the outputs are the amount of each pollutant's emission. The outcome of this supervised learning approach is a model that can best describe the relationship between inertial sensor data and the level of emission of the given construction equipment. In a supervised learning method, inputs and associated outputs (i.e., ground truth) are provided to the model in the training phase. Learning algorithms use training data to produce inferred functions, which are then used to map new examples using a test dataset that has not been seen by the model during training [76]. In this research, the training portions constituted 70% of the dataset and the rest was kept out for testing.

In order to identify the best model, classic ML algorithms have been used to see if they are capable of predicting emission levels directly from accelerometer and gyroscope sensors data. Four learning algorithms were developed for this study: Neural Network (NN), Decision Tree Regression (DTR), Random Forest (RF), and Linear Regression (LR). Each algorithm is described briefly in the following subsections.

#### 3.3.1. Linear Regression (LR)

LR is the simplest of all among the algorithms tested in this research. Nevertheless, its simplicity sometimes results in acceptable results calculated in a fast computation process. In LR, the model assumes that input variables ($x$) and output variables ($y$) are linearly related. Precisely, the LR model calculates $y$ from a linear combination of the input variables $x$'s. The literature from statistics often refers to LR as multiple linear regression when there are multiple input variables such as the case in this research [77]. This method of modeling corresponds explanatory variables to a scalar response with a linear approach. LR, similar to all regression analysis methods, targets conditional probability distributions of response, rather than their joint probability distributions as is the case with multivariate analysis (i.e., where there is more than one dependent variable). Due to the nature of the data and the complexity of relationships between the variables in this study, it is not expected that the LR model leads to an acceptable result. However, LR models have been developed in this research for all three pollutants to verify this hypothesis.

#### 3.3.2. Neural Network (NN)

NN is commonly used to solve a wide variety of scientific and engineering problems [78]. A two-layer NN includes only one input layer, one hidden layer, and one output layer. A standard multilayer perceptron NN is comprised of an input layer, a hidden layer, and an output layer, and each layer consists of nodes. The number of layers and nodes (i.e., NN architecture) are important factors in determining the performance of the model. In this research, Gradient Descent was used as the optimization algorithm. Gradient Descent is one of the most commonly used optimization algorithms for NN models and is used heavily in both linear regression and classification problems [79]. Gradient Descent depends on the first-order derivative of the loss function. The function calculates how the weights should be altered to reach the minimum loss. By backpropagation, the loss is passed from one layer to another in the model and its parameters, or weights, are modified based on the loss for the best result [80]. The proposed model in this research uses the learning rate of 0.01 which is a hyperparameter that affects the speed of learning by changing the model to a degree each time its weights are updated. In addition, the Sigmoid function was used as an activation function to set all values in the input data to a value between 0 and 1.

### 3.3.3. Decision Tree Regression (DTR)

DTR is a ML method commonly used to predict discretely or continuous values in classification or regression problems. Decision trees are based on tree structures, which build regression or classification models. A decision tree is developed incrementally as a dataset is broken down into ever-smaller subsets. As a result, a tree is created with decision nodes and leaf nodes. Building decision trees is based on the ID3 algorithm introduced by Quinlan; ID3 uses a top-down, greedy search through the space of possible branches without backtracking [81]. DTR has different variables that affect the results. Different values have been assigned to these variables in this research to enable finding the best answer. Finding the best answer among all the tested models with different values for these variables are discussed in Section 4.

### 3.3.4. Random Forest (RF)

RF is the combination of several decision trees. For regression problems, the mean or average prediction of the individual trees is calculated as the answer. As such, RF usually outperforms the decision tree. Although RF increases bias in a single tree, it generally reduces the variance. This results in a higher accuracy because a more complex or flexible model will typically have a higher variance and bias due to overfitting. The model, however, predicts the target variable more accurately when averaged over several predictions. While an underfit or oversimplified model has lower variance, it will likely be more biased since it lacks the tools to capture trends in the data [82]. Thus, the number of trees is an important variable in RF. RF is used in this study with the different number of trees for each gas to obtain the best answer. The other parameters are set according to the results for DRT.

### 3.3.5. Performance Metrics

Generally, ML models can be used for classification or regression problems. In classification, the goal is to separate the data into multiple categorical groups through the discovery of a model or function. In regression, a function or model is developed to assist in separating the data into continuous real values, rather than using discrete classes or values [83]. Due to the continuous space of the problem in this research, regression models are used to predict the amount of each pollutant released. The accuracy of the models is determined by performance metrics that show how well a model can predict the results. For regression models, there are four most commonly used performance metrics: coefficient of determination ($R^2$), root-mean-square error (*RMSE*), mean absolute error (*MAE*), and normalized root-mean-square error (*NRMSE*) as shown in Equations 1-4.

$$R^2 = 1 - \frac{\sum(y - \hat{y})^2}{\sum(y - \bar{y})^2} \quad (1)$$

$$RMSE = \sqrt{\frac{1}{n}\sum(y - \hat{y})^2} \quad (2)$$

$$MAE = \frac{1}{n}\sum |y - \hat{y}| \quad (3)$$

$$NRMSE = \frac{RMSE}{y_{max} - y_{min}} * 100 \quad (4)$$

where $y$ is the predicted value, $\hat{y}$ is the actual value or ground truth, $n$ is the number of examples used to determine the accuracy, and $y_{min}$ and $y_{max}$ are the minimum and maximum predicted values in the set.

## 4. Results

Each learning algorithm includes several factors such as hyperparameters, model architecture, and constituting algorithms the choice of which has a great effect on the performance of the model. Therefore, in this section, various variables are adjusted and the performance of the models are compared for each algorithm to determine the best algorithm and architecture. However, before outlining the model performance pertaining to each algorithm, a what-if analysis was performed to ensure that the use of both sensors versus only using accelerometers as well as the additional step of feature extraction result in an improved performance.

*4.1 LR*

As indicated before, LR is not expected to produce superb results. All other learning methods that are tested in this study use nonlinear functions that better suit the nature of the data and relationships between the independent and dependent variables used in this research. Table 3 shows the subpar performance of LR according to all four performance measures.

*Table 3. The performance metrics analyzed by linear regression.*

| Emitted Gas | $R^2$ | RMSE | MAE | NRMSE |
|---|---|---|---|---|
| CO | 0.06 | 45.12 | 32.12 | 16.40 |
| NO | 0.19 | 84.42 | 71.39 | 20.84 |
| $NO_2$ | 0.33 | 4.92 | 3.65 | 9.85 |
| $NO_x$ | 0.18 | 85.35 | 72.26 | 19.44 |
| $CO_2$ | 0.22 | 113.98 | 92.25 | 18.65 |

*4.2. NN*

The results of the different NN architectures are shown in Table 4. In this Table, the number of nodes in each hidden layer for each model are shown in the Layer Architecture column in square brackets (the number of nodes in the input layer is equal to the number of features used and the output layer has only one node). On average, the $R^2$ value for the NN models to estimate the *CO* gas produced by the excavator is 0.805. In other words, the NN model indicates that 80.5% of the variation in the *CO* emission for this equipment can be explained by the data captured by the accelerometer and gyroscope sensors attached to the equipment. Table 1 also shows the other performance metrics identified in the previous section. Results indicate that more layers and more nodes did not improve the performance of the model by a significant margin. Therefore, model architecture with three hidden layers of 100, 90, and 80 nodes which resulted in a slightly better performance than the others was selected as the NN model to compare with the other learning algorithms.

*Table 4. Performance metrics related to ANN analysis for CO emission.*

| Layer Architecture | $R^2$ | RMSE | MAE | NRMSE |
|---|---|---|---|---|
| [40 30] | 0.81 | 20.72 | 12.41 | 7.55 |
| [40 30 20] | 0.80 | 20.77 | 12.43 | 7.56 |
| [100 90 80] | 0.82 | 19.85 | 11.79 | 7.22 |
| [200 190 180] | 0.79 | 20.88 | 12.36 | 7.59 |

The two other emission types investigated in this study are $NO_x$ and $CO_2$. A similar process was adopted to examine the effect of model architecture on the prediction output. Results are presented in Tables 5 and 6. As an example, for $CO_2$ a visualized representation of the models is also provided in Figure 2.

*Table 5. Performance metrics related to ANN analysis for NOx emission.*

| Layer Architecture | $R^2$ | RMSE | MAE | NRMSE |
|---|---|---|---|---|
| [40 30] | 0.62 | 58.97 | 43.17 | 13.73 |
| [40 30 20] | 0.64 | 56.48 | 39.80 | 12.86 |
| [100 90 80] | 0.65 | 55.92 | 39.47 | 12.74 |
| [200 190 180] | 0.65 | 55.82 | 38.67 | 12.71 |

*Table 6. Performance metrics related to ANN analysis for $CO_2$ emission.*

| Layer Architecture | $R^2$ | RMSE | MAE | NRMSE |
|---|---|---|---|---|
| [40 30] | 0.71 | 69.94 | 45.78 | 11.45 |
| [40 30 20] | 0.71 | 69.94 | 45.78 | 11.45 |

| [100 90 80]    | 0.78 | 60.51 | 37.32 | 9.90 |
| [200 190 180]  | 0.77 | 60.61 | 36.46 | 9.92 |

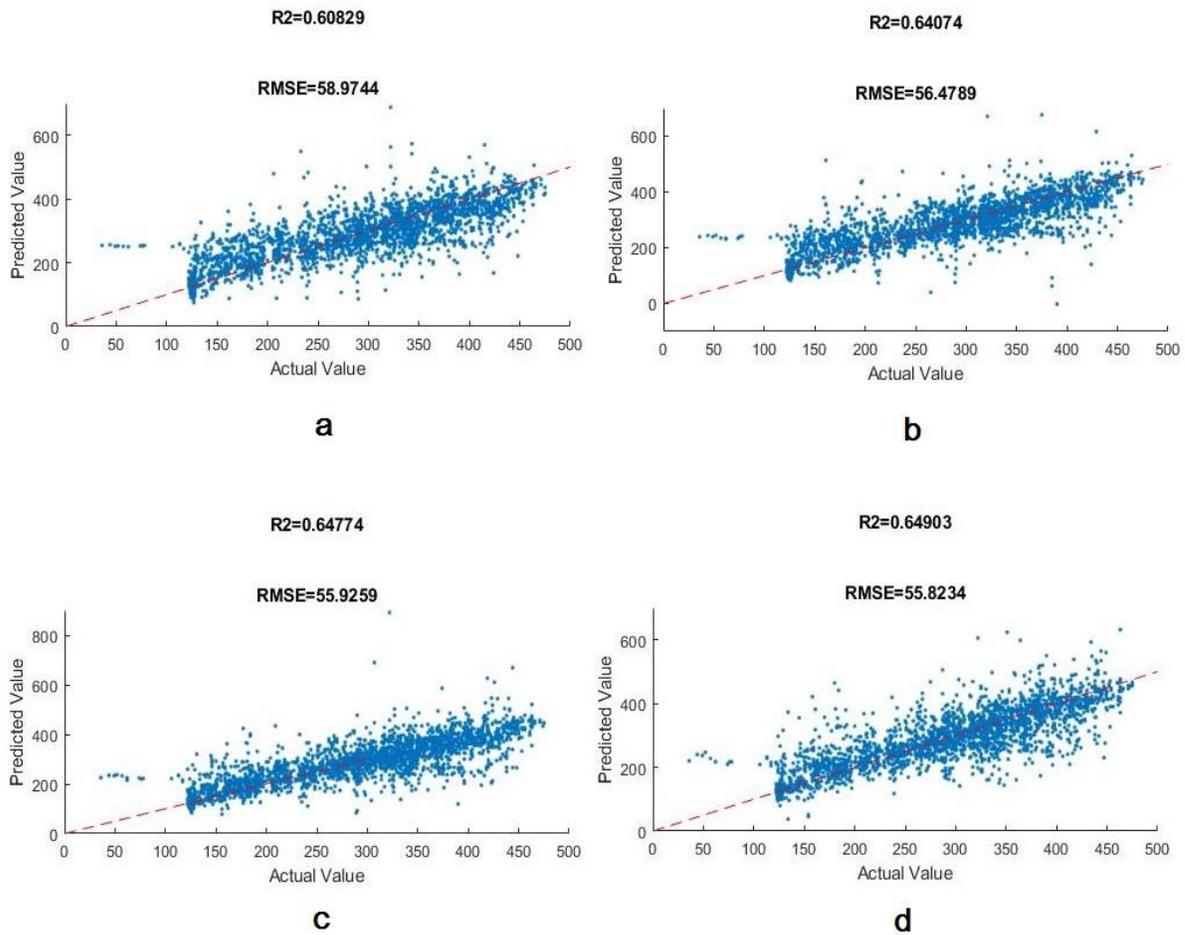

*Figure 2. The $CO_2$ regression lines are trained by the neural network. (a) ANN with Two layers, 40 and 30 neurons. (b) ANN with Three layers, 40, 30, and 20 neurons. (c) ANN with Three layers, 100, 90, and 80 neurons. (d) ANN with Three layers, 200, 190, and 180 neurons.*

As shown in Table 6 for the $NO_x$, there is not much difference in the results of the models with 200 and 100 nodes in the first hidden layer and the metric factors converge after 100 neurons (the $R^2$ of 0.65 is considered for further comparison). For the $CO_2$ the best architecture is the same as the $CO$ and the $R^2$ is 0.78. The performance of the best model for all the pollutants is presented in Table 7 where the results of the $NO$ and $NO_2$ are included for comparison.

*Table 7. The best performance factors for all gasses.*

| **Emitted Gas** | $R^2$ | RMSE | MAE | NRMSE |
| **CO** | 0.82 | 19.85 | 11.79 | 7.22 |
| **NO** | 0.66 | 54.71 | 38.02 | 13.51 |
| **NO$_2$** | 0.78 | 2.85 | 1.65 | 5.70 |
| **NO$_x$** | 0.65 | 55.82 | 38.67 | 12.71 |
| **CO$_2$** | 0.78 | 60.51 | 37.32 | 9.90 |

### 4.3. DTR

In the DTR, there are variables such as the maximum number of decision splits, the minimum number of leaf node observations (*minLeafSize*), and the minimum number of branch node observations (*minParentSize*) that control the depth of the tree and thus, affect its performance. The maximum number of decision splits determines the maximum number of the nodes in each branch and the optimum result is obtained when this variable is equal to the number of training data minus one. The two variables that determine the accuracy of the model are *minLeafSize* and *minParentSize*.

Figure 3 shows the regression lines developed by adjusting these variables in the DTR model. In Figure 3(a) *minLeafSize* is equal to 1 and *minParentSize* is 5. With a fix *minParentSize*, the increase in *minLeafSize* results in lower $R^2$ and the higher value of different errors (i.e., RMSE, MAE, and NRMSE) after *minLeafSize equal to two*. So, the optimum number for minLeafSize is two. There is not any significant growth in the accuracy when the minParentSize increase to ten. Consequently, the optimum state for the decision tree is represented in Figure. 3(d) with performance factors: $R^2 = 0.85$, $RMSE = 17.69$, $MAE = 6.78$, $NRMSE = 6.43$. Also, Table 10 shows the best results obtained by DTR through changing its configuration. The optimum setting for the *NO* and *NO$_2$* is assumed as same as *NO$_X$*.

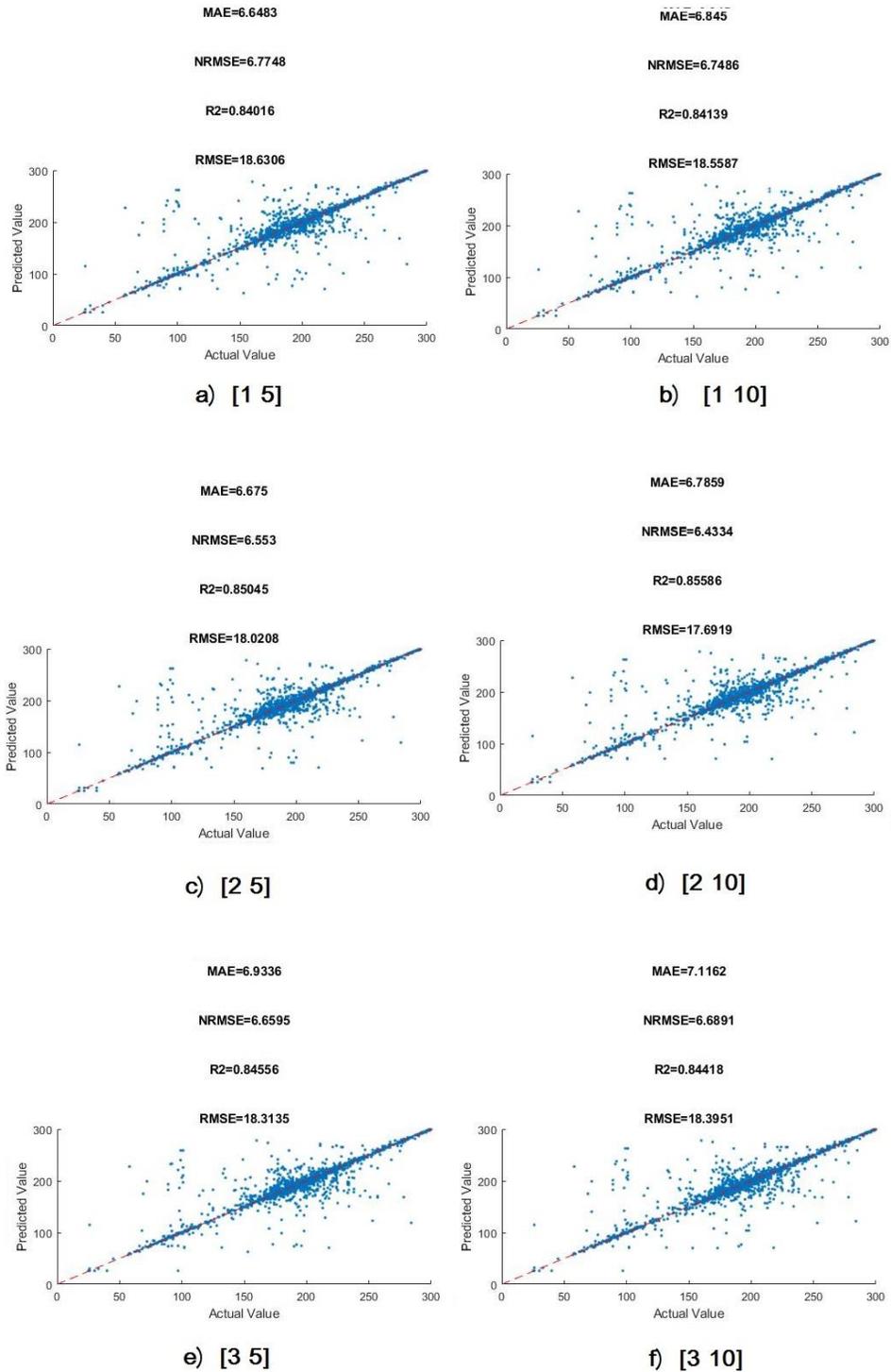

*Figure. 3. The CO regression lines analyzed by decision tree algorithm. (a) MinLeafSize=1, MinParentSize=5. (b) MinLeafSize=1, MinParentSize=10. (c) MinLeafSize=2, MinParentSize=5. (d) MinLeafSize=2, MinParentSize=10. (e) MinLeafSize=5, MinParentSize=5. (f) MinLeafSize=3, MinParentSize=10.*

*Table 8. Performance metrics related to Decision Tree analysis for $NO_x$ emission.*

| [MinLeafSize MinParentSize] | $R^2$ | RMSE | MAE | NRMSE |
|---|---|---|---|---|
| [1 5] | 0.77 | 44.46 | 19.63 | 10.12 |
| [1 10] | 0.78 | 43.98 | 20.42 | 10.02 |
| [2 5] | 0.76 | 45.82 | 20.92 | 10.43 |
| [2 10] | 0.77 | 45.08 | 21.16 | 10.26 |
| [3 5] | 0.76 | 45.22 | 21.30 | 10.30 |
| [3 10] | 0.77 | 44.87 | 21.47 | 10.22 |

*Table 9. Performance metrics related to Decision Tree analysis for $CO_2$ emission.*

| [MinLeafSize MinParentSize] | $R^2$ | RMSE | MAE | NRMSE |
|---|---|---|---|---|
| [1 5] | 0.86 | 47.52 | 18.00 | 7.77 |
| [1 10] | 0.86 | 47.28 | 18.53 | 7.73 |
| [2 5] | 0.87 | 46.97 | 18.17 | 7.67 |
| [2 10] | 0.86 | 47.45 | 19.05 | 7.76 |
| [3 5] | 0.86 | 48.02 | 19.46 | 7.86 |
| [3 10] | 0.86 | 47.79 | 19.77 | 7.82 |

*Table 10. Performance metrics related to DTR.*

| Gas | [MinLeafSize MinParentSize] | $R^2$ | RMSE | MAE | NRMSE |
|---|---|---|---|---|---|
| CO | [2 10] | 0.85 | 117.69 | 6.78 | 6.43 |
| NO | [1 10] | 0.79 | 42.39 | 19.66 | 10.47 |
| NO2 | [1 10] | 0.87 | 2.15 | 0.74 | 4.31 |
| NOx | [1 10] | 0.78 | 43.98 | 20.42 | 10.02 |
| CO2 | [2 5] | 0.87 | 46.97 | 18.17 | 7.67 |

*4.4. RF*

The most important feature in RF is minimum samples at each leaf (*minLeafSize*) and the number of trees. *minLeafSize* was set according to the best answer in the decision tree for each pollutant. The accuracy is expected to increase as the number of trees goes up. Figure 4 plots $R^2$ against the number of trees for the RF model to identify the convergence point from which this performance metric remains constant as the number of trees goes up.

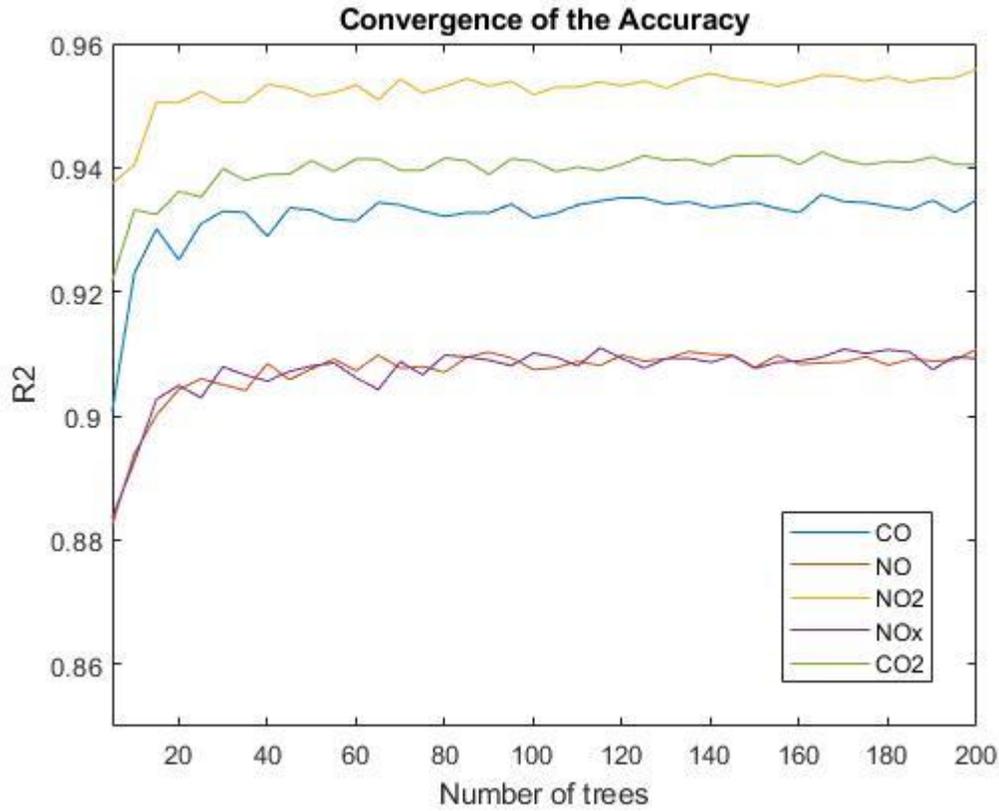

*Figure 4. Amount of coefficient of determination in different trees for random forest*

Results of this analysis indicate that the best performance factors are obtained when the number of trees is 150, 85, and 90 for *CO*, *NO$_X$*, and *CO$_2$* respectively. All the performance metrics for the optimum number of trees are presented in Table 11.

*Table 11. Performance metrics related to RF.*

| Gas | Number of Trees | $R^2$ | RMSE | MAE | NRMSE |
|---|---|---|---|---|---|
| **CO** | 150 | 0.94 | 11.70 | 5.48 | 4.25 |
| **NO** | 135 | 0.91 | 28.35 | 16.77 | 7.00 |
| **NO₂** | 200 | 0.95 | 1.30 | 0.64 | 2.60 |
| **NOx** | 85 | 0.91 | 28.19 | 16.74 | 6.42 |
| **CO₂** | 90 | 0.94 | 31.63 | 15.98 | 5.17 |

## 5. Discussion

In this section, first the performance of the four algorithms developed in this study are compared using the model that in each case manifested the best outcome in terms of the metrics used as the benchmark. Following this comparison, the results of previous studies in the field of emission prediction are evaluated to contextualize the findings of this study within the literature.

*5.1 Models Performance and Pollutants Predictability*

An analysis of the calculated $R^2$ metrics indicates that RF outperformed the three other algorithms used in this research in predicting the emission levels for all three pollutants. It was 0.94, 0.91, and 0.94 for *CO*, *NO$_x$*, *CO$_2$*, respectively. RF was followed by DT. Therefore, it can be concluded that with an appropriate number of trees, RF can be a superb model when the equipment emission levels are explained by inertial sensor data. NN ranks three and shows a lower performance compared to RF and DT. Table 12 shows the $R^2$ values for the best models in each case.

*Table 12. Best coefficient of determination calculated by different algorithms*

| $R^2$ | NN | DTR | RF | Linear Regression |
|---|---|---|---|---|
| **CO** | 0.82 | 0.85 | 0.94 | 0.06 |
| **NO$_X$** | 0.65 | 0.77 | 0.91 | 0.18 |
| **CO₂** | 0.78 | 0.87 | 0.94 | 0.22 |

According to Table 11, the RF model performed similarly in the case of *CO* and *CO$_2$* and resulted in a very high R$_2$ value. A close examination of *NO$_X$* prediction (which followed these two pollutants very closely) indicates that the $R^2$ metric for *NO$_2$* was, in fact, equal to that of *CO* and *CO$_2$* whereas this value for *NO* was significantly lower in the case of all four algorithms (see Tables 3,7,10 and 11). Plotting the collected raw emission data against time (Figure 5) reveals that *NO* exhibits a different pattern in terms of the number and range of fluctuations per time unit compared to other pollutants. While this different pattern does not explain the lower performance, it does indicate the difference that exists in the nature of the collected data that may call for different learning methodologies.

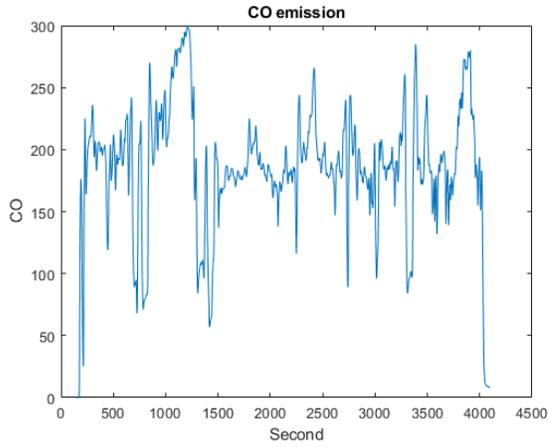
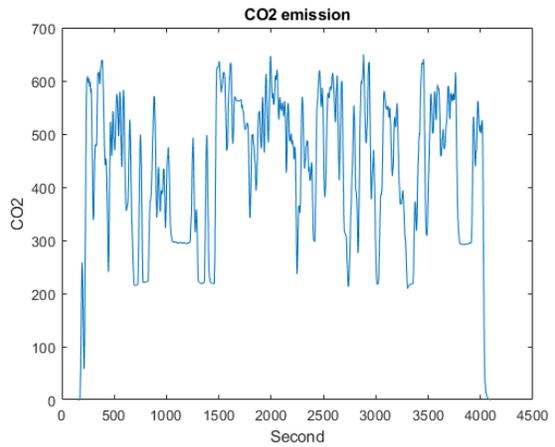
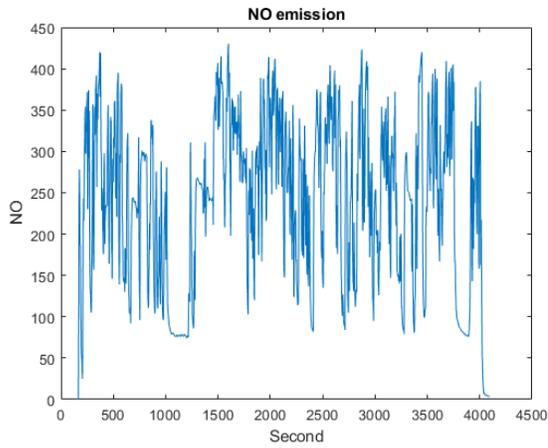
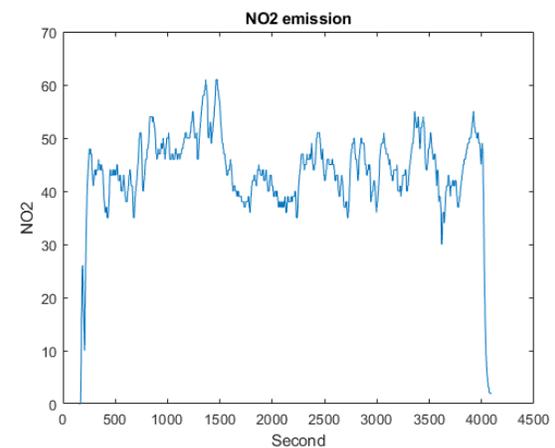
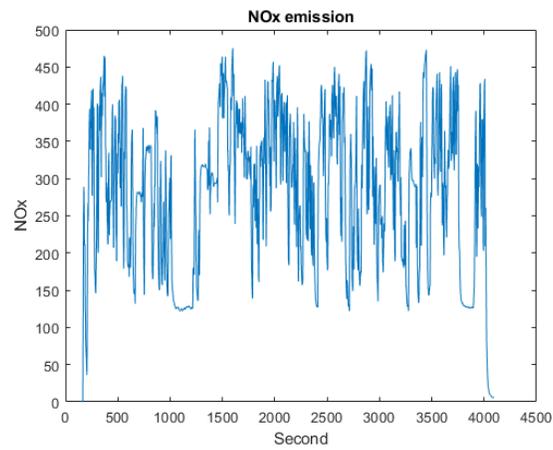

*Figure 5. Amount of the emissions in the second*

5.2 Evaluation of the Results in the Context of Previous Studies

Several studies have aimed at forecasting the emission of combustion engines in transportation vehicles. Most of the previous studies in this area of research targeted on-road vehicles. The presented paper is the first study with the objective of predicting emissions of the construction equipment only through the evaluation of vehicle movement and without collecting data related to factors such as engine parameters, weather conditions, speed of the equipment, or fuel consumption. Therefore, the purpose of outlining the results of similar work in the literature in this section is not a point-to-point competition between the methodologies or numerical values of the results between this research and those studies. Rather, it is to contextualize the outcome of this research within a similar work and compare the results of predicting emission for heavy equipment versus on-road vehicles. The coefficient of determination or $R^2$ is used as the comparison metric since it is a common metric in most of the previous similar studies.

Xu et al. (2020) developed an XGBoost model to predict $CO_2$ and $PM_{2.5}$ where they examined a variety of variables on trip-level emission including meteorology, trip characteristics (such as time of day), driving characteristics (such as idling frequency), and driver characteristics (such as experience in driving). They obtained an $R^2$ value of 0.84 [84]. A complete ensemble empirical mode decomposition with adaptive noise (CEEMDAN) and a long short-term memory (LSTM) neural network is proposed for estimating transient $NO_X$ emissions by Yu et al. (2021). They used engine attributes variables such as vehicle speed, engine speed, torque percentage, instantaneous fuel consumption, accelerator pedal opening as the input of the CEEMDAN-LSTM model where they obtained an $R^2$ of 0.98 [37]. Barati and Shen (2016) developed a multivariate linear regression (MLR) where engine attributes, operational parameters, environmental factors, and fuel type were the factors used as the input of the model for on-road construction equipment emissions. The $R^2$ values for the MLR method were 0.96, 0.89, 0.93 and 0.90 for $CO_2$, $CO$, $NO_x$, $HC$, respectively [50]. Table 12 summarizes these results along with the final $R^2$ values obtained in this study for comparison.

*Table 12. Comparison between different algorithms.*

| $R^2$ | Random Forest (This Study) | MLR [50] | CEEMDAN-LSTM [37] | CEEMDAN-XGBoost [37] | XGBoost [84] |
|---|---|---|---|---|---|
| **CO** | 0.94 | 0.89 | - | - | - |
| **NOx** | 0.91 | 0.93 | 0.98 | 0.96 | - |
| **CO₂** | 0.94 | 0.96 | - | - | 0.84 |

## 6. Conclusion

The objective of this study was to examine the possibility of using and IoT-ML integrated framework to collect and transmit multi-modal data in order to predict construction equipment emissions. The study examined the relationship between construction equipment movements and emission rates. This shows a strong bond between the two of them. This study laid a foundation for managing the level of greenhouse gases produced by construction equipment. A case study of an excavator was selected to predict the amount of emissions by kinetic sensors installed in the cabin and arm. Several supervised learning algorithms were employed to ML models for predicting emissions from construction equipment. The results show that random forest by the coefficient of determination ($R^2$) equal to 0.94, 0.91 and 0.94 respectively for *CO*, $NO_X$, and $CO_2$ is the best algorithm.

Different supervised learning algorithms were employed in this research to predict the emissions from construction equipment just by data extracted from accelerometer and gyroscope sensors without any information about the engine attributes such as load and size, environmental factors such as temperature, air pressure and slope, fuel type and payload. According to the outputs, there is a correlation between the inputs and the actual emissions and it is possible to estimate the exhaust gasses just by two smart mobile phones in various projects. Therefore, it is not necessary to mount expensive devices on equipment such as PEMSs and engine parameter recorders.

As a pioneering attempt to estimate heavy equipment emission using inertial sensors, this study can inspire future research in this area. Nevertheless, there are a few limitations that can be addressed in future research. First, this research deploys traditional shallow learning algorithms, while the recent advancements in the area of deep learning can prove valuable to enhance the results of similar studies. Computational models with multiple layers of processing can learn abstractions of data through deep learning. In future studies, deep learning algorithms such as LSTM or convolutional neural network (CNN) can replace traditional methods. They may represent better performance metrics. Second, even in the realm of traditional ML methods, the performance of other supervised algorithms such as Naive Bayes, perceptron, relevance vector machine (RVM), and support vector machine (SVM) can be assessed. Third, a more comprehensive research project can focus on several pollutants including but not limited to the ones explored in this study. Finally, collecting data from more than one piece of equipment and exploring subject-independent models where the models are trained on one machine but are tested on another can enhance the generalizability and application of the developed methodologies and algorithms.

**Acknowledgements**

The presented work is supported by the California Senate Bill 1 (SB1): California State University Transportation Consortium (CSUTC), grant #1852. The authors gratefully acknowledge CSUTC for their financial support. The authors would also like to thank Kiewit Corporation for allowing access to their job sites for data collection. Any opinions, findings, conclusions, and recommendations expressed in this paper are those of the authors and do not necessarily represent those of the CSUTC or Kiewit Corporation.